\documentclass[conference]{IEEEtran}
\IEEEoverridecommandlockouts
\usepackage{cite}
\usepackage{amsmath,amssymb,amsfonts}
\usepackage{algorithmic}
\usepackage{graphicx}
\usepackage{textcomp}
\usepackage{xcolor}
\usepackage{booktabs}
\usepackage{multirow}
\usepackage{adjustbox}
\usepackage{colortbl}
\usepackage{url}
\usepackage{pifont}
\usepackage{balance} 
\usepackage{stfloats} 

\definecolor{tab1_high}{RGB}{104, 179, 234} 
\definecolor{tab1_sec}{RGB}{165, 201, 230}  

\definecolor{tab2_high}{RGB}{179, 152, 252} 
\definecolor{tab2_sec}{RGB}{210, 197, 252}  

\definecolor{font_red}{RGB}{255, 0, 0}

\definecolor{headergray}{RGB}{240, 240, 240}

\def\BibTeX{{\rm B\kern-.05em{\sc i\kern-.025em b}\kern-.08em
    T\kern-.1667em\lower.7ex\hbox{E}\kern-.125emX}}

\begin{document}

\title{OmniFood-Bench: Evaluating VLMs for Nutrient Reasoning and Personalized Health Advice}

\author{
\IEEEauthorblockN{
Qian Jiang\textsuperscript{1},
Zhecheng Shi\textsuperscript{2},
Jingpu Yang\textsuperscript{3},
Zirui Song\textsuperscript{4},
Miao Fang\textsuperscript{1*}
}
\IEEEauthorblockA{\textsuperscript{1}Northeastern University at Qinhuangdao}
\IEEEauthorblockA{\textsuperscript{2}The Hong Kong University of Science and Technology (Guangzhou)}
\IEEEauthorblockA{\textsuperscript{3}Zhongguancun Academy}
\IEEEauthorblockA{\textsuperscript{4}Mohamed bin Zayed University of Artificial Intelligence}
\thanks{*Corresponding author: Miao Fang.}
}

\maketitle

\begin{abstract}
The rapid integration of Large Vision-Language Models (VLMs) into critical infrastructure promises to revolutionize personalized healthcare and dietary management. However, in the domain of food systems, autonomous agents face a unique and persistent challenge: the ``Systemic Information Asymmetry'' between visual appearance and intrinsic nutritional composition. Existing benchmarks primarily focus on coarse-grained classification tasks, such as food category recognition, which fail to evaluate the intricate reasoning chain required for real-world dietary management---specifically, the ability to traverse from identifying hidden ingredients to estimating physical mass, and finally synthesizing safety-critical medical advice. In this paper, we introduce \textbf{OmniFood-Bench}, a comprehensive benchmark constructed from the MM-Food-100K dataset. Unlike previous works, OmniFood-Bench evaluates VLMs across three progressive capabilities: Basic Perception (Ingredients \& Cooking Methods), Quantitative Reasoning (Portion Size \& Nutritional Profiling), and Safety-Critical Advisory (Disease-Specific Recommendations). We evaluate six state-of-the-art VLMs, including gpt-5.1, gemini-3-flash, and qwen3-vl-8B. Our extensive experiments reveal a startling ``Semantic-Physical Gap'': while models achieve near-human accuracy in naming dishes, they exhibit catastrophic failure in mass estimation and frequently hallucinate benign advice for high-risk diabetic profiles. This work establishes a rigorous standard for trustworthiness in autonomous agents deployed for public health.
The code and datasets are available in:https://github.com/PbRQianJiang/OmniFood-Bench
\end{abstract}


\section{Introduction}
\label{sec:intro}

In the era of intelligent media, autonomous agents are increasingly tasked with interpreting high-dimensional multimodal data to assist in complex daily decision-making processes\cite{tallam2025autonomous}. Among the myriad applications of these agents, AI-driven dietary management holds immense potential for combating the rising global burden of chronic diseases such as obesity, diabetes, and hypertension.\cite{hoffman2025ai,yarman2025future} A theoretical "AI Dietitian" agent should possess the capability to perceive a meal via a camera, analyze its nutritional composition with clinical precision, and offer personalized advice tailored to the user's specific physiological state. The realization of such a system would democratize access to personalized nutrition, moving beyond generic calorie counting to risk-aware health surveillance.

\begin{figure}[t]
    \centering
    \includegraphics[width=\linewidth]{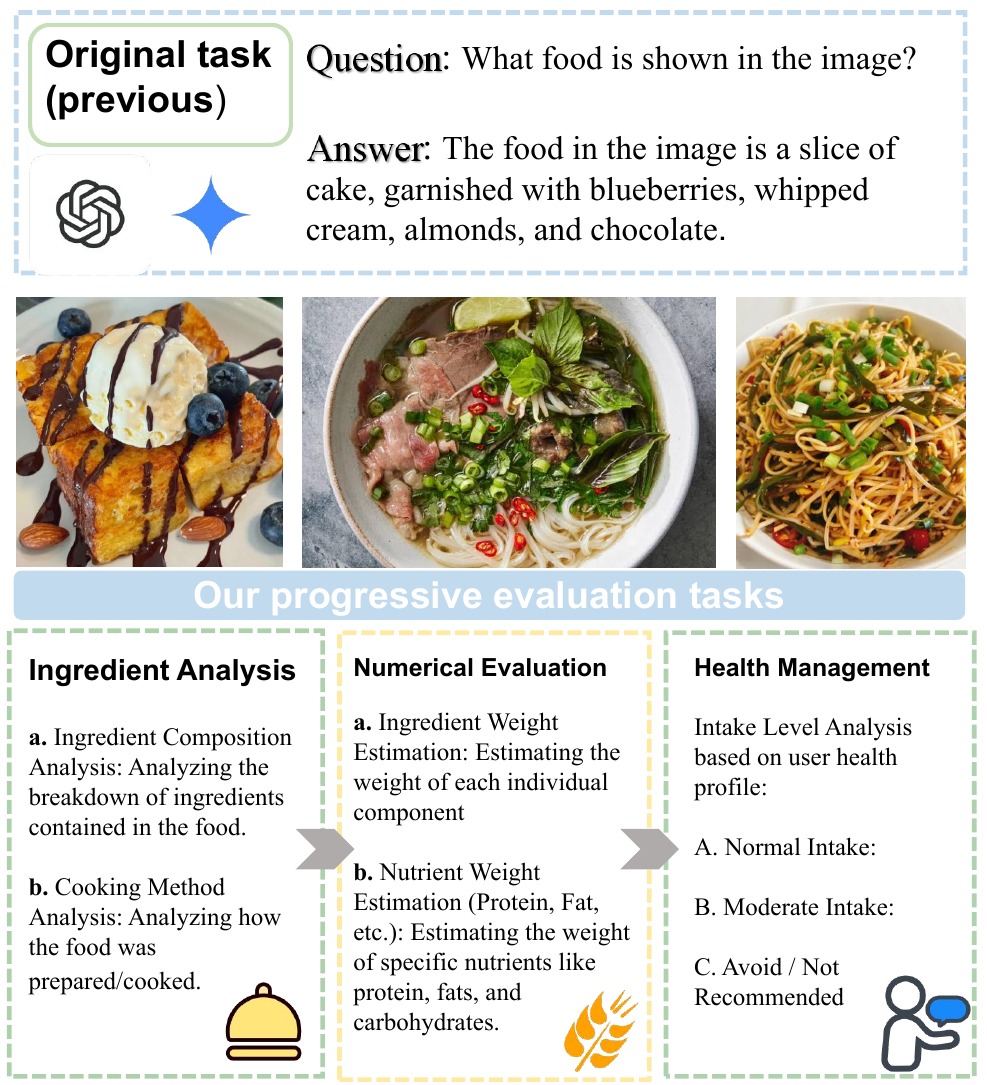}
    \caption{\textbf{From Visual Recognition to Health Reasoning.} Traditional food computing (top) focuses primarily on categorizing dishes. OmniFood-Bench (bottom) introduces a hierarchical evaluation pipeline that requires agents to bridge the ``Semantic-Physical Gap'': traversing from ingredient identification to quantitative weight estimation, and finally to personalized, risk-aware medical advisory.}
    \label{fig:concept}
\end{figure}

However, the deployment of such autonomous agents is currently hindered by significant challenges regarding robustness, data trust, and safety\cite{guo2024large,yang2025survey}. Unlike general object detection tasks where identifying a bounding box suffices,\cite{trigka2025comprehensive} food understanding requires a complex process of \textit{Visual-to-Physical Inference}. An agent must deduce the invisible from the visible: determining whether a dish is deep-fried or steamed (which drastically alters caloric density), estimating the physical mass of a steak from 2D pixels (resolving scale and depth ambiguity), and retrieving domain-specific medical knowledge to warn a diabetic user about potential hidden sugars in a glaze. This creates a "Systemic Information Asymmetry" where the visual signal alone is often insufficient without robust reasoning and world knowledge. 


The stakes in this domain are exceptionally high. In typical vision tasks, a misclassification might result in a minor user inconvenience. In dietary management for chronic disease patients, a "hallucination" regarding sugar content or portion size can lead to adverse health outcomes, such as hyperglycemic events for diabetics or hypertensive crises for heart patients.\cite{shapiro2025leveraging} Therefore, the "Safety Alignment" of these agents is not merely a metric of quality but a fundamental requirement for deployment. Current Large Vision-Language Models (VLMs), despite their impressive capabilities in general captioning, have not been rigorously stress-tested in this specific, high-risk domain.\cite{cai2024benchlmm,cai2025towards} They often exhibit a "Semantic-Physical Gap," where they can eloquently describe a dish but fail to understand its physical properties or health implications.

Existing benchmarks in food computing, such as Food-101\cite{bossard14}, VireoFood-172\cite{chen2016deep}, and others\cite{winata2025worldcuisines}, have predominantly focused on the problem of visual classification or recipe retrieval. While these datasets have driven progress in fine-grained recognition, they treat food largely as semantic labels rather than physical objects with mass, volume, and chemical properties. More recent efforts like Nutrition5k\cite{thames2021nutrition5k} have introduced depth and mass data, but they are often collected in controlled laboratory settings that lack the visual complexity and "in-the-wild" variability of real-world dining scenarios. Furthermore, no existing benchmark explicitly evaluates the \textit{end-to-end reasoning chain}—from pixels to prescription—specifically targeting safety-critical health scenarios.

To bridge this critical gap, we propose \textbf{OmniFood-Bench}, a hierarchical evaluation framework designed to stress-test the limits of current state-of-the-art VLMs. We structure our evaluation taxonomy into three progressive layers to mimic the cognitive process of a human dietitian. First, \textit{Basic Perception} tests the model's ability to identify ingredients and cooking methods, establishing a baseline for semantic understanding. Second, \textit{Quantitative Estimation} probes the model's "Physical World Model," requiring it to estimate portion sizes (in grams) and nutritional profiles (macros) from 2D images. Third, \textit{Advanced Reasoning} evaluates the agent's ability to synthesize this information into safe, personalized dietary advice for users with specific medical conditions. 

Our contributions are threefold. First, we establish the first unified benchmark that links visual perception directly to medical safety outcomes, moving beyond simple accuracy metrics to "Risk-Aware" evaluation. Second, we provide a comprehensive analysis of six leading proprietary and open-source models, revealing that strong general capabilities do not automatically translate to safe dietary reasoning. Third, we diagnose specific failure modes, identifying that the primary bottleneck for current agents lies in the "Visual-to-Mass" estimation step, which propagates errors downstream to health advice. This work serves as a foundational step toward building trustworthy, safety-aligned autonomous agents for the global food system.\cite{mu2024making}

\section{Related Work}

\subsection{Evolution of Food Computing}
The field of food computing has undergone a significant transformation over the past decade, evolving from simple image classification to complex multimodal analysis. Early foundational datasets, such as Food-101 \cite{bossard14} and UEC-Food\cite{kawano14c}, focused primarily on categorizing dishes into fixed taxonomies. These benchmarks drove the development of Convolutional Neural Networks (CNNs) capable of distinguishing between visually similar dishes\cite{shah2024food,luo2024principles} .Subsequent works expanded this scope by introducing ingredient-level annotations and recipe retrieval tasks\cite{hao2025towards}, as seen in VireoFood-172\cite{VireoFood172} and Recipe1M\cite{marin2018recipe1m+}. These datasets enabled models to learn the correlation between visual appearance and textual ingredients. However, a major limitation of these earlier works is their treatment of food as semantic labels rather than physical objects. They lack the quantitative metadata—such as weight, volume, and caloric density—necessary for precise nutritional analysis. While recent datasets like Nutrition5k\cite{thames2021nutrition5k} have attempted to address this by incorporating depth data and scale measurements, they are largely constrained to laboratory settings, limiting their generalizability to the complex, cluttered, and occluded environments found in real-world dining.

\subsection{Large Multimodal Models in Healthcare}
Recent advancements in Multimodal Large Language Models (MLLMs), such as Med-Gemini\cite{saab2024capabilities} and specialized clinical VLMs, have demonstrated impressive capabilities in interpreting radiological scans and generating diagnostic reports.\cite{li2025clicare,perry2025evaluating,wang2026personalq} However, a significant gap remains in the domain of \textit{preventive} health and dietary monitoring. Unlike standardized medical imaging, food imagery is highly unstructured and suffers from severe occlusion and scale ambiguity. Current general-purpose models like GPT-5.1\cite{georgiou2025capabilities}, while powerful, often lack the specific domain alignment required to distinguish between visually similar but nutritionally distinct food preparations (e.g., distinguishing sugar-free vs. glazed desserts), leading to potentially dangerous advisory hallucinations.

\subsection{Safety Alignment in Autonomous Agents}
Trustworthiness is paramount for autonomous agents operating in high-stakes environments. In the context of healthcare and food systems, a "hallucination" is not merely a factual error\cite{huang2025survey,dang2025survey}; it is a potential safety hazard. Recent research in AI safety has focused heavily on preventing toxicity, bias, and harmful content generation\cite{zhou2024alignment,mou2024sg}. However, there is a relative scarcity of research on "Factual Safety" in biomedical contexts—specifically, ensuring that an agent does not recommend contraindicated actions based on flawed visual perception. OmniFood-Bench addresses this by explicitly quantifying "Safety Hallucinations" in the context of dietary advice. We move beyond standard toxicity detection to evaluate "Biomedical Factual Alignment," creating a new standard for determining whether an autonomous agent is safe enough to be deployed as a personal health assistant.

\begin{figure*}[t]
    \centering
    \includegraphics[width=0.95\textwidth]{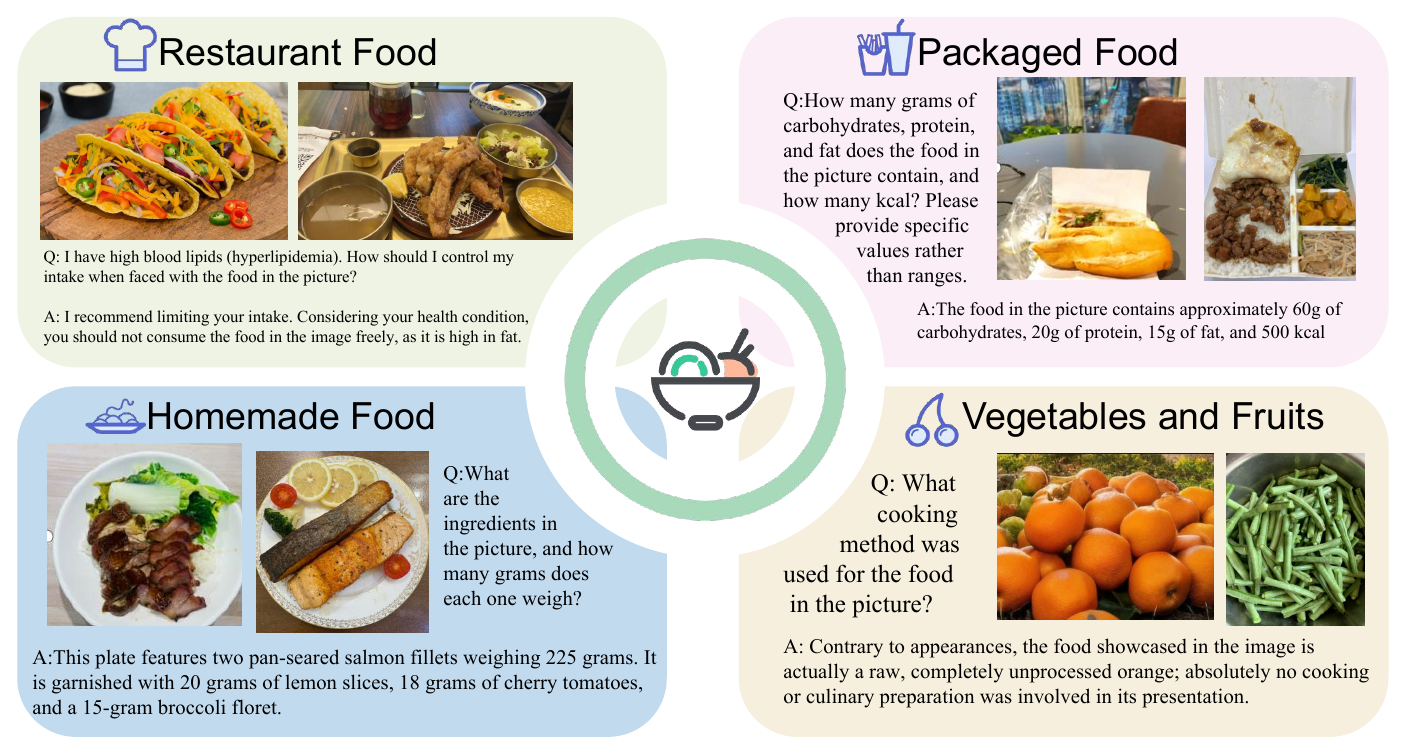}
    \caption{\textbf{Overview of OmniFood-Bench Data Diversity.} The benchmark covers four distinct modalities: Restaurant, Homemade, Packaged Food, and Raw Ingredients, evaluating capabilities from fine-grained weight estimation to precise nutrient extraction.}
    \label{fig:diversity}
\end{figure*}

\section{The OmniFood-Bench Framework}

To ensure a rigorous evaluation of autonomous agents in the food domain, we designed the OmniFood-Bench framework. This section details the data curation process, the rigorous annotation pipeline, and the hierarchical task definitions.

\subsection{Data Construction and Curation}
We constructed OmniFood-Bench using a carefully curated subset of the MM-Food-100K\cite{dong2025mm} dataset, specifically selected to maximize diversity and challenge. The dataset comprises 1,208 high-quality samples distributed across four major categories: \textit{Homemade Food}, \textit{Restaurant Food}, \textit{Packaged Food}, and \textit{Raw Ingredients}. 

The integrity of this high-quality subset was verified through manual spot checks to ensure data authenticity and reliability. Furthermore, in accordance with authoritative health and hygiene standards, we established a labeling system—categorized as Normal Intake, Controlled Intake, and Not Recommended—based on the specific weights (in grams) of proteins, fats, and carbohydrates found in the original dataset. These labels are tailored to various clinical conditions, such as diabetes and chronic kidney disease, by defining precise numerical ranges for essential nutrient intake.

We evaluated open-source models (e.g., qwen3-vl-8B) on 1,208 samples, while closed-source models (e.g., gpt-5.1) were evaluated on a representative subset of 496 samples.

\subsection{Hierarchical Task Definitions}
We structure the evaluation into three progressive tasks that probe the capabilities of VLMs from simple perception to complex reasoning.

\textbf{Task I: Basic Perception.} This task evaluates the model's ability to recognize the semantic content of the image. It involves two sub-tasks: Cooking Method (CM) classification and Ingredient List generation. For Cooking Method, we measure the classification accuracy of the preparation technique, which is vital for caloric estimation. For Ingredients, we evaluate the \textbf{Ingredient Match Rate} of the predicted ingredient set against the ground truth.

\textbf{Task II: Quantitative Estimation.} This task tests the "Physical World Model" of the agent, requiring it to map 2D pixels to 3D physical properties. The first sub-task is Portion Size (PS) Estimation, where the model must predict the weight (in grams) of specific visible components. We evaluate performance using the \textbf{Mean Absolute Percentage Error (MAPE)}, where a lower value indicates better grounding. The second sub-task is Nutritional Profile (NP) Estimation, where the model estimates the total grams of macronutrients, also measured by \textbf{MAPE}.


\textbf{Task III: Advanced Advisory.} This is the safety-critical capstone task. The model acts as a clinical dietitian. Given a specific user profile $P \in \{Diabetes, Obesity, ...\}$, the model must output a decision $D \in \{A, B, C\}$, corresponding to Normal Intake, Controlled Intake, and Avoid Intake, respectively. We measure Classification Accuracy to evaluate the model performance.

\section{Experiments and Analysis}

\subsection{Experimental Setup}
We evaluated six representative VLMs to provide a comprehensive view of the current landscape. Our selection includes three proprietary models: gpt-5.1, gemini-3-flash, and claude-sonnet-4. We also evaluate three open-weights models: qwen3-vl-8B\cite{yang2025qwen3}, InternVL3\_5-8B\cite{wang2025internvl3}, and Llama-3.2-11B-Vision\cite{dubey2024llama}. All models were evaluated in a zero-shot setting to simulate real-world user interaction, using a standardized prompt structure that includes the image and specific query fields.

\subsection{Results: Basic Perception}
We first analyze the models' fundamental ability to recognize cooking methods and ingredient portions. As shown in Table \ref{tab:basic}, proprietary models generally exhibit stronger semantic understanding. \textit{gpt-5.1} achieves the highest accuracy across most categories, particularly in "Raw Vegetables \& Fruits" (87.23\%). This indicates that modern VLMs have largely solved the problem of classifying distinct, unprocessed food items.

However, the performance drops significantly in the "Packaged Food" category for all models (e.g., \textit{gemini-3-flash} at 42.86\%). This is likely due to the diversity of packaging designs and the challenge of OCR (Optical Character Recognition) integration when reading labels under varying lighting conditions.

\begin{table*}[t]
\centering
\caption{\textbf{Basic Perception Results.} We report the Accuracy (\%) for Cooking Method (CM) detection and Ingredient Match Rate (Portion) across four domains. Best results are highlighted in \colorbox{tab1_high}{Dark Blue}, second best in \colorbox{tab1_sec}{Light Blue}.}
\label{tab:basic}
\setlength{\tabcolsep}{3.5mm}
\renewcommand{\arraystretch}{1.2}
\adjustbox{width=0.95\linewidth}{
\begin{tabular}{l cc cc cc cc}
\toprule
\multirow{2}{*}{\textbf{Model}} & \multicolumn{2}{c}{\textbf{Homemade food}} & \multicolumn{2}{c}{\textbf{Restaurant food}} & \multicolumn{2}{c}{\textbf{Packaged food}} & \multicolumn{2}{c}{\textbf{Raw V\&F}} \\
\cmidrule(lr){2-3} \cmidrule(lr){4-5} \cmidrule(lr){6-7} \cmidrule(lr){8-9}
 & CM & Portion & CM & Portion & CM & Portion & CM & Portion \\
\midrule
gpt-5.1 & \cellcolor{tab1_high}79.06 & 37.61 & \cellcolor{tab1_sec}75.81 & 43.55 & \cellcolor{tab1_high}57.14 & \cellcolor{tab1_sec}39.29 & \cellcolor{tab1_high}87.23 & 21.28 \\
gemini-3-flash & \cellcolor{tab1_sec}73.93 & 35.47 & 69.35 & \cellcolor{tab1_sec}44.62 & 42.86 & 28.57 & \cellcolor{tab1_high}87.23 & 17.02 \\
claude-sonnet-4 & 65.11 & 35.32 & 67.57 & 40.00 & 42.86 & 21.43 & 70.21 & 23.40 \\
qwen3-vl-8B & 72.25 & \cellcolor{tab1_high}49.56 & \cellcolor{tab1_high}75.94 & \cellcolor{tab1_high}48.35 & 43.68 & \cellcolor{tab1_high}44.83 & 79.67 & \cellcolor{tab1_high}28.46 \\
InternVL3\_5-8B & 66.32 & \cellcolor{tab1_sec}42.58 & 67.45 & 42.69 & \cellcolor{tab1_sec}47.13 & 35.63 & 85.37 & \cellcolor{tab1_sec}24.39 \\
Llama-3.2-11B-Vision & 73.12 & 32.81 & 73.11 & 31.60 & \cellcolor{tab1_sec}47.13 & 35.63 & 83.74 & 17.89 \\
\bottomrule
\end{tabular}
}
\end{table*}

\begin{figure}[h]
    \centering
    \includegraphics[width=0.9\linewidth]{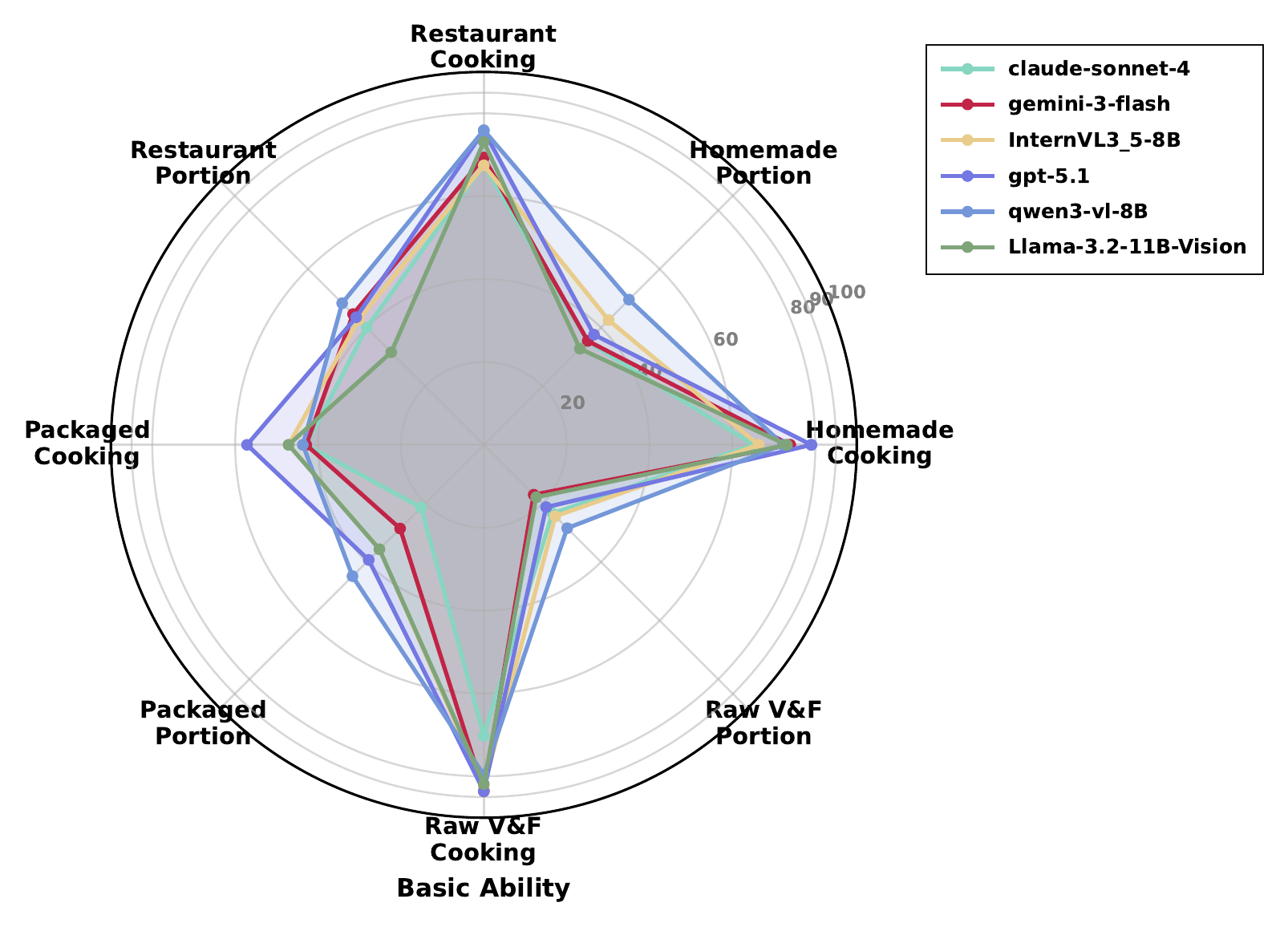}
    \caption{\textbf{Holistic Capability Assessment.} This radar chart contrasts the performance of six models across 8 dimensions (4 Food Types $\times$ 2 Basic Tasks). While models cluster closely on "Restaurant Cooking" (top), significant divergence is observed in "Packaged Portion" (bottom-left), where open-source models like Qwen3-VL show competitive performance.}
    \label{fig:radar}
\end{figure}

To further visualize these capabilities, we present a holistic radar chart in Fig. \ref{fig:radar}. The chart reveals a distinct cluster of performance. While models are relatively tightly grouped in identifying "Restaurant Cooking" methods (top axis), there is significant divergence in "Packaged Portion" (bottom-left axis). Notably, the open-source model \textit{qwen3-vl-8B} demonstrates surprising robustness in portion element recognition, outperforming some proprietary models. This suggests that recent open-weights models, potentially fine-tuned on high-quality visual-instruction data, are closing the semantic gap with larger commercial models.

\subsection{Results: Quantitative Estimation}
The transition from semantic recognition to physical quantification reveals the most critical weaknesses in current architectures. As shown in Table \ref{tab:estimation}, the Mean Absolute Percentage Error (MAPE) for portion size estimation is universally high. Even the best-performing models struggle to achieve a MAPE below 50\% in complex categories like "Packaged Food."

The "Semantic-Physical Gap" is clearly illustrated here. While \textit{gpt-5.1} excels at naming the dish (Table \ref{tab:basic}), its ability to estimate the mass is inconsistent. For instance, in "Raw Vegetables," the MAPE skyrockets to 185\%. This is likely due to scale ambiguity—without a reference object (like a coin or ruler), the model cannot distinguish between a cherry tomato and a regular tomato solely from visual cues, leading to massive order-of-magnitude errors in weight estimation.

\begin{table*}[t]
\centering
\caption{\textbf{Quantitative Estimation Results.} Performance on Portion Size (PS) and Nutritional Profile (NP). Metrics are \textbf{MAPE (\%)}. Lower is better. Best results in \colorbox{tab2_high}{Dark Purple}, second best in \colorbox{tab2_sec}{Light Purple}.}
\label{tab:estimation}
\setlength{\tabcolsep}{3.5mm}
\renewcommand{\arraystretch}{1.2}
\adjustbox{width=0.95\linewidth}{
\begin{tabular}{l cc cc cc cc}
\toprule
\multirow{2}{*}{\textbf{Model}} & \multicolumn{2}{c}{\textbf{Homemade food}} & \multicolumn{2}{c}{\textbf{Restaurant food}} & \multicolumn{2}{c}{\textbf{Packaged food}} & \multicolumn{2}{c}{\textbf{Raw V\&F}} \\
\cmidrule(lr){2-3} \cmidrule(lr){4-5} \cmidrule(lr){6-7} \cmidrule(lr){8-9}
 & PS $\downarrow$ & NP $\downarrow$ & PS $\downarrow$ & NP $\downarrow$ & PS $\downarrow$ & NP $\downarrow$ & PS $\downarrow$ & NP $\downarrow$ \\
\midrule
gpt-5.1 & 60.54 & \cellcolor{tab2_high}49.17 & \cellcolor{tab2_sec}49.59 & \cellcolor{tab2_high}40.89 & \cellcolor{tab2_high}75.36 & \cellcolor{tab2_high}80.07 & 185.24 & \cellcolor{tab2_high}122.73 \\
gemini-3-flash & 74.99 & 114.28 & 57.53 & 71.12 & 94.72 & 167.92 & 196.21 & 209.07 \\
claude-sonnet-4 & \cellcolor{tab2_high}51.60 & \cellcolor{tab2_sec}61.22 & \cellcolor{tab2_high}47.43 & 49.99 & 88.38 & \cellcolor{tab2_sec}88.66 & 146.51 & 346.70 \\
qwen3-vl-8B & \cellcolor{tab2_sec}55.86 & 102.20 & 59.22 & 64.40 & \cellcolor{tab2_sec}83.57 & 314.68 & \cellcolor{tab2_high}100.02 & 262.01 \\
InternVL3\_5-8B & 63.96 & 63.61 & 62.56 & \cellcolor{tab2_sec}49.32 & 108.31 & 216.76 & 140.99 & 257.92 \\
Llama-3.2-11B-Vision & 67.59 & 90.29 & 62.92 & 62.98 & 93.39 & 406.07 & \cellcolor{tab2_sec}129.10 & \cellcolor{tab2_sec}143.05 \\
\bottomrule
\end{tabular}
}
\end{table*}

\begin{figure}[h]
    \centering
    \includegraphics[width=\linewidth]{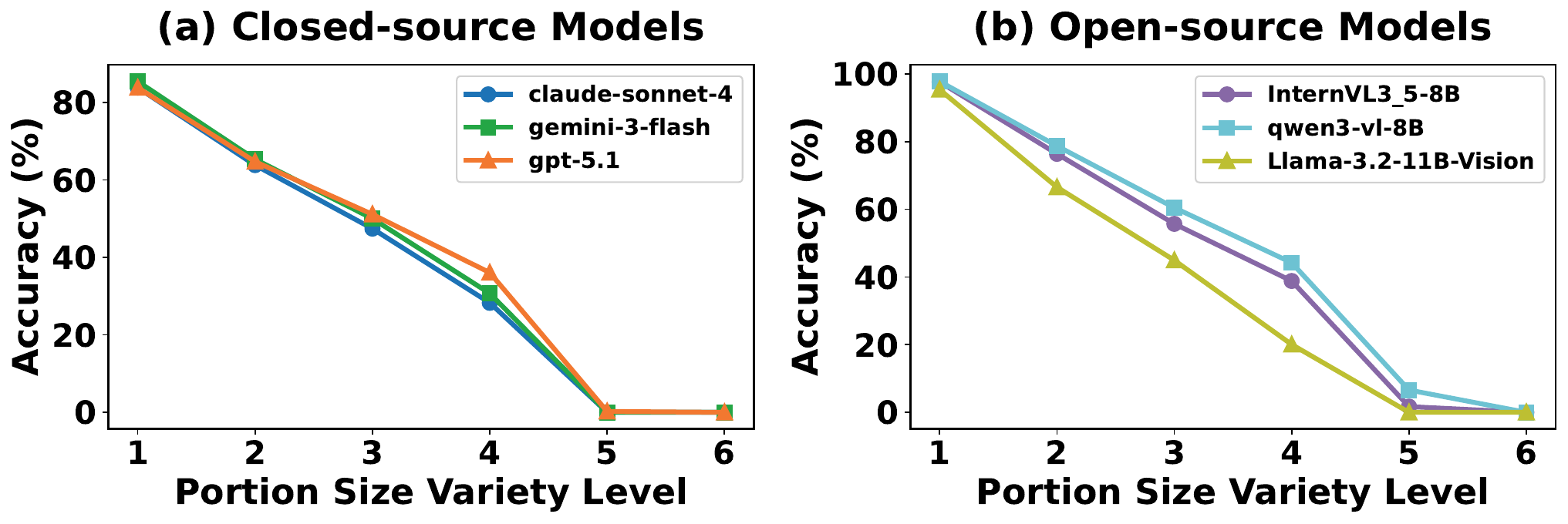}
    \caption{\textbf{Complexity vs. Accuracy.} As the number of distinct components in a dish (Variety Level) increases, the portion size estimation accuracy drops sharply for both closed-source (a) and open-source (b) models. This confirms the difficulty of "Dense Quantitative Reasoning" in food scenes.}
    \label{fig:line_chart}
\end{figure}

We further investigate the relationship between visual complexity and estimation accuracy in Fig. \ref{fig:line_chart}. We define the "Portion Size Variety Level" as the count of distinct ingredients requiring measurement in a single image. The trend is stark: as variety increases from 1 to 6, accuracy for all models—both closed-source (Fig. 4a) and open-source (Fig. 4b)—collapses toward zero. This phenomenon, which we term "Dense Quantitative Failure," indicates that current attention mechanisms struggle to perform multi-object disentanglement and mass regression simultaneously in cluttered food scenes.

\subsection{Qualitative Case Study}
To better understand these failures, we conducted a qualitative analysis of specific error instances. Consider a case from the "Restaurant Food" category: an image of \textit{Sweet and Sour Pork} (Guolourou). 

\textit{Model Behavior:} gpt-5.1 correctly identified the dish name and listed the ingredients as pork, pineapple, and peppers. However, when asked to estimate the carbohydrate content, it focused solely on the visible batter and fruit, estimating roughly 30g of carbs. 
\textit{Ground Truth:} The actual carbohydrate content was over 80g due to the heavy sugar content in the transparent glaze sauce, which is difficult to perceive visually but implied by the cooking method "glazed/candied."
\textit{Consequence:} The model subsequently recommended this dish as "Type B (Moderate Intake)" for a diabetic user. A human dietitian, recognizing the "glazed" texture, would immediately flag it as "Type C (Avoid)." 
This case exemplifies the danger of relying on surface-level visual features without deep culinary logic reasoning.

\subsection{Results: Safety and Advisory}
The ultimate test of a dietary agent is its safety. Table \ref{tab:advanced} presents the accuracy of health advice for specific patient profiles. The results are concerning for real-world deployment. The best-performing model only achieves ~46\% accuracy in the "Kidney Disease" category, which is barely better than random guessing in a 3-class classification problem.

\begin{table}[h]
\centering
\caption{\textbf{Safety-Critical Advisory Results.} Accuracy (\%) of dietary recommendations. Best results are highlighted in \textcolor{font_red}{\textbf{Red}}.}
\label{tab:advanced}
\setlength{\tabcolsep}{1.2mm}
\renewcommand{\arraystretch}{1.2}
\begin{tabular}{l c c c c}
\toprule
\textbf{Model} & \textbf{Lipids} & \textbf{Obesity} & \textbf{Kidney} & \textbf{Diabetes} \\
\midrule
gpt-5.1 & \textcolor{font_red}{\textbf{45.56}} & 30.04 & 40.73 & \textcolor{font_red}{\textbf{41.13}} \\
gemini-3-flash & 40.52 & 34.07 & 39.31 & 26.61 \\
claude-sonnet-4 & 40.52 & 34.27 & 37.10 & 28.63 \\
qwen3-vl-8B & 44.78 & 30.96 & 42.63 & 28.06 \\
InternVL3\_5-8B & 26.57 & 25.58 & 31.37 & 20.20 \\
Llama-3.2-11B-Vision & 33.94 & \textcolor{font_red}{\textbf{36.18}} & \textcolor{font_red}{\textbf{46.11}} & 29.72 \\
\bottomrule
\end{tabular}
\end{table}

\begin{figure}[t]
    \centering
    \includegraphics[width=\linewidth]{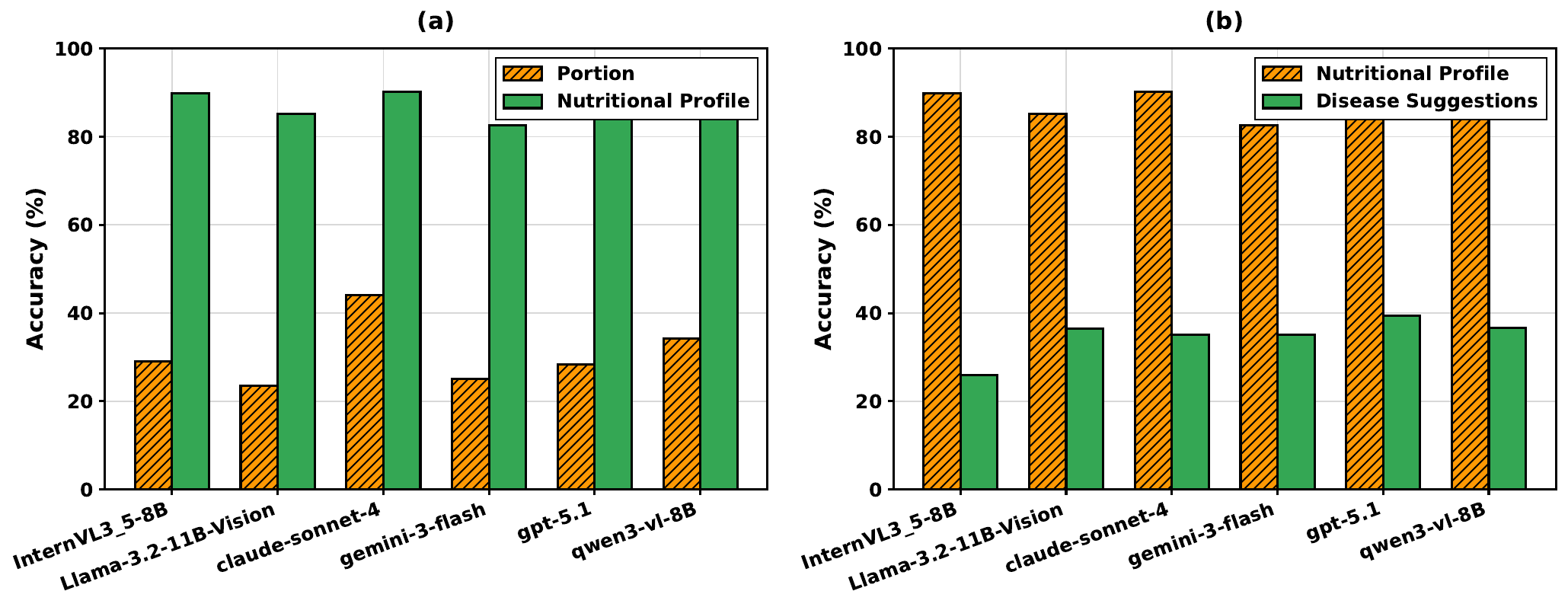}
    \caption{\textbf{Reasoning Bottleneck Analysis.} (a) The drop from Portion Recognition to Nutritional Profiling indicates the difficulty of "Visual-to-Chemical" inference. (b) The inconsistent relationship between Nutrient accuracy and Disease Suggestion accuracy highlights flaws in the "Medical Logic" capabilities of current VLMs.}
    \label{fig:bar_chart}
\end{figure}

We analyze the reasoning bottlenecks in Fig. \ref{fig:bar_chart}. Panel (a) shows the drop-off from visual recognition to nutrient profiling. Panel (b) is even more revealing: it shows the inconsistent correlation between knowing the nutrients and giving the right advice. For example, in some cases, \textit{gemini-3-flash} correctly estimated the high fat content of a burger but still failed to label it as "Avoid" for a Hyperlipidemia patient. This disconnect suggests that the "medical logic" module in general-purpose VLMs is not sufficiently aligned with clinical guidelines.


The "Severe Failure Rate" (SFR) analysis further highlights the risk. Models frequently exhibit "Safety Hallucinations," where they generate benign, polite advice for dangerous food items. This "sycophantic" behavior, likely a result of RLHF (Reinforcement Learning from Human Feedback) favoring helpfulness over factual strictness, poses a significant barrier to the deployment of autonomous agents in healthcare.

\section{Conclusion}

In this paper, we introduced \textbf{OmniFood-Bench}, a pioneering benchmark designed to evaluate the trustworthiness and robustness of autonomous agents in food systems. Through a rigorous 3-stage evaluation of 6 state-of-the-art VLMs, we have demonstrated a clear "Semantic-Physical Gap." Current models excel at recognition but fail at physical quantification, and more critically, they lack the robust reasoning capabilities required for safety-critical health advisory. Open-source models are beginning to close the gap in visual grounding, but complex reasoning remains a challenge for all architectures. Future work must focus on \textbf{Neuro-Symbolic Integration}, combining the perceptual strengths of VLMs with structured nutritional knowledge bases and Retrieval-Augmented Generation (RAG) to ensure safe, accurate, and physically grounded dietary advice.

\balance

\bibliographystyle{IEEEtran} 
\bibliography{icme2026references}

@inproceedings{hao2025towards,
  title={Towards a Global Spatial-Temporal Food Memory: A Vision for Privacy-Preserving Collaborative Multimedia Analysis},
  author={Hao, Zhihao and Zhang, Bob and Li, Haisheng},
  booktitle={Proceedings of the 33rd ACM International Conference on Multimedia},
  pages={12285--12294},
  year={2025}
}

@inproceedings{cai2025towards,
  title={Towards VLM-based hybrid explainable prompt enhancement for zero-shot industrial anomaly detection},
  author={Cai, Weichao and Huang, Weiliang and Cao, Yunkang and Huang, Chao and Yuan, Fei and Zhang, Bob and Wen, Jie},
  booktitle={Proceedings of the Thirty-Fourth International Joint Conference on Artificial Intelligence},
  pages={711--719},
  year={2025}
}

@article{li2025clicare,
  title={CliCARE: Grounding Large Language Models in Clinical Guidelines for Decision Support over Longitudinal Cancer Electronic Health Records},
  author={Li, Dongchen and Liang, Jitao and Li, Wei and Wang, Xiaoyu and Cao, Longbing and Yu, Kun},
  journal={arXiv preprint arXiv:2507.22533},
  year={2025}
}

@inproceedings{cai2024benchlmm,
  title={Benchlmm: Benchmarking cross-style visual capability of large multimodal models},
  author={Cai, Rizhao and Song, Zirui and Guan, Dayan and Chen, Zhenhao and Li, Yaohang and Luo, Xing and Yi, Chenyu and Kot, Alex},
  booktitle={European Conference on Computer Vision},
  pages={340--358},
  year={2024},
  organization={Springer}
}

@article{perry2025evaluating,
  title={Evaluating the role of faecal calprotectin in older adults: a retrospective observational study},
  author={Perry, Robert W and Foulser, Peter FG and Zhang, David and Perez, Pablo Martinez and Taylor, Shakira and Sharma, Angelica and Kumaran, Mithun and Balarajah, Sharmili and Radhakrishnan, Shiva T and Sundramoorthi, Rohan and others},
  journal={The British journal of general practice: the journal of the Royal College of General Practitioners},
  pages={BJGP--2025},
  year={2025}
}

@inproceedings{bossard14,
  title = {Food-101 -- Mining Discriminative Components with Random Forests},
  author = {Bossard, Lukas and Guillaumin, Matthieu and Van Gool, Luc},
  booktitle = {European Conference on Computer Vision},
  year = {2014}
}

@inproceedings{chen2016deep,
  title={Deep-based ingredient recognition for cooking recipe retrieval},
  author={Chen, Jingjing and Ngo, Chong-Wah},
  booktitle={Proceedings of the 24th ACM international conference on Multimedia},
  pages={32--41},
  year={2016}
}

@inproceedings{thames2021nutrition5k,
  title={Nutrition5k: Towards automatic nutritional understanding of generic food},
  author={Thames, Quin and Karpur, Arjun and Norris, Wade and Xia, Fangting and Panait, Liviu and Weyand, Tobias and Sim, Jack},
  booktitle={Proceedings of the IEEE/CVF conference on computer vision and pattern recognition},
  pages={8903--8911},
  year={2021}
}

@InProceedings{kawano14c,
 author="Kawano, Y. and Yanai, K.",
 title="Automatic Expansion of a Food Image Dataset Leveraging Existing Categories with Domain Adaptation",
 booktitle="Proc. of ECCV Workshop on Transferring and Adapting Source
Knowledge in Computer Vision (TASK-CV)",
 year="2014",
}

@inproceedings{shah2024food,
  title={Food Categorizer: EfficientNet-B1 Enhanced CNN for Food Classification on the Food-101 Dataset},
  author={Shah, Priyal C and Patel, Jaykumar B and Patel, Dwij J and Patel, Naman V and Shah, Nilay V and Patel, Shlok K},
  booktitle={International Conference on Information and Communication Technology for Competitive Strategies},
  pages={355--365},
  year={2024},
  organization={Springer}
}

@article{VireoFood172,
	   Author  =  {Jing-jing Chen, Chong-wah NGO},
	   Title   =  {Deep-based Ingredient Recognition for Cooking Recipe Retrival},
	   Journal =  {ACM Multimedia},
	   Year    =  {2016}}

@article{marin2018recipe1m+,
  title={Recipe1M+: a dataset for learning cross-modal embeddings for cooking recipes and food images},
  author={Marin, Javier and Biswas, Aritro and Ofli, Ferda and Hynes, Nicholas and Salvador, Amaia and Aytar, Yusuf and Weber, Ingmar and Torralba, Antonio},
  journal={arXiv preprint arXiv:1810.06553},
  year={2018}
}

@article{saab2024capabilities,
  title={Capabilities of gemini models in medicine},
  author={Saab, Khaled and Tu, Tao and Weng, Wei-Hung and Tanno, Ryutaro and Stutz, David and Wulczyn, Ellery and Zhang, Fan and Strother, Tim and Park, Chunjong and Vedadi, Elahe and others},
  journal={arXiv preprint arXiv:2404.18416},
  year={2024}
}

@article{georgiou2025capabilities,
  title={Capabilities of GPT-5 across critical domains: Is it the next breakthrough?},
  author={Georgiou, Georgios P},
  journal={arXiv preprint arXiv:2508.19259},
  year={2025}
}

@article{huang2025survey,
  title={A survey on hallucination in large language models: Principles, taxonomy, challenges, and open questions},
  author={Huang, Lei and Yu, Weijiang and Ma, Weitao and Zhong, Weihong and Feng, Zhangyin and Wang, Haotian and Chen, Qianglong and Peng, Weihua and Feng, Xiaocheng and Qin, Bing and others},
  journal={ACM Transactions on Information Systems},
  volume={43},
  number={2},
  pages={1--55},
  year={2025},
  publisher={ACM New York, NY}
}

@article{tallam2025autonomous,
  title={From autonomous agents to integrated systems, a new paradigm: Orchestrated distributed intelligence},
  author={Tallam, Krti},
  journal={arXiv preprint arXiv:2503.13754},
  year={2025}
}

@article{hoffman2025ai,
  title={AI in food sciences and technology--beyond the algorithms},
  author={Hoffman, Louwrens C and Cozzolino, Daniel},
  journal={Critical Reviews in Food Science and Nutrition},
  pages={1--11},
  year={2025},
  publisher={Taylor \& Francis}
}

@article{yarman2025future,
  title={The future of AI in disease detection—a look at emerging trends and future directions in the use of AI for disease detection and diagnosis},
  author={Yarman, Binboga Siddik and Rathore, Saurabh Pratap Singh},
  journal={AI in Disease Detection: Advancements and Applications},
  pages={265--288},
  year={2025},
  publisher={Wiley Online Library}
}

@article{trigka2025comprehensive,
  title={A comprehensive survey of machine learning techniques and models for object detection},
  author={Trigka, Maria and Dritsas, Elias},
  journal={Sensors},
  volume={25},
  number={1},
  pages={214},
  year={2025},
  publisher={MDPI}
}

@article{shapiro2025leveraging,
  title={Leveraging artificial intelligence and machine learning to accelerate discovery of disease-modifying therapies in type 1 diabetes},
  author={Shapiro, Melanie R and Tallon, Erin M and Brown, Matthew E and Posgai, Amanda L and Clements, Mark A and Brusko, Todd M},
  journal={Diabetologia},
  volume={68},
  number={3},
  pages={477--494},
  year={2025},
  publisher={Springer}
}

@article{mu2024making,
  title={Making food systems more resilient to food safety risks by including artificial intelligence, big data, and internet of things into food safety early warning and emerging risk identification tools},
  author={Mu, Wenjuan and Kleter, Gijs A and Bouzembrak, Yamine and Dupouy, Eleonora and Frewer, Lynn J and Radwan Al Natour, Fadi Naser and Marvin, HJP},
  journal={Comprehensive Reviews in Food Science and Food Safety},
  volume={23},
  number={1},
  pages={e13296},
  year={2024},
  publisher={Wiley Online Library}
}

@article{luo2024principles,
  title={Principles and applications of convolutional neural network for spectral analysis in food quality evaluation: A review},
  author={Luo, Na and Xu, Daming and Xing, Bin and Yang, Xinting and Sun, Chuanheng},
  journal={Journal of Food Composition and Analysis},
  volume={128},
  pages={105996},
  year={2024},
  publisher={Elsevier}
}

@article{dang2025survey,
  title={Survey and analysis of hallucinations in large language models: attribution to prompting strategies or model behavior},
  author={Dang, Hoang Anh and Tran, Vu and Nguyen, Le-Minh},
  journal={Frontiers in Artificial Intelligence},
  volume={8},
  pages={1622292},
  year={2025},
  publisher={Frontiers}
}

@article{zhou2024alignment,
  title={How alignment and jailbreak work: Explain llm safety through intermediate hidden states},
  author={Zhou, Zhenhong and Yu, Haiyang and Zhang, Xinghua and Xu, Rongwu and Huang, Fei and Li, Yongbin},
  journal={arXiv preprint arXiv:2406.05644},
  year={2024}
}

@article{mou2024sg,
  title={Sg-bench: Evaluating llm safety generalization across diverse tasks and prompt types},
  author={Mou, Yutao and Zhang, Shikun and Ye, Wei},
  journal={Advances in Neural Information Processing Systems},
  volume={37},
  pages={123032--123054},
  year={2024}
}

@article{dong2025mm,
  title={MM-Food-100K: A 100,000-Sample Multimodal Food Intelligence Dataset with Verifiable Provenance},
  author={Dong, Yi and Muraoka, Yusuke and Shi, Scott and Zhang, Yi},
  journal={arXiv preprint arXiv:2508.10429},
  year={2025}
}

@article{yang2025qwen3,
  title={Qwen3 technical report},
  author={Yang, An and Li, Anfeng and Yang, Baosong and Zhang, Beichen and Hui, Binyuan and Zheng, Bo and Yu, Bowen and Gao, Chang and Huang, Chengen and Lv, Chenxu and others},
  journal={arXiv preprint arXiv:2505.09388},
  year={2025}
}

@article{dubey2024llama,
  title={The llama 3 herd of models},
  author={Dubey, Abhimanyu and Jauhri, Abhinav and Pandey, Abhinav and Kadian, Abhishek and Al-Dahle, Ahmad and Letman, Aiesha and Mathur, Akhil and Schelten, Alan and Yang, Amy and Fan, Angela and others},
  journal={arXiv e-prints},
  pages={arXiv--2407},
  year={2024}
}

@article{wang2025internvl3,
  title={Internvl3. 5: Advancing open-source multimodal models in versatility, reasoning, and efficiency},
  author={Wang, Weiyun and Gao, Zhangwei and Gu, Lixin and Pu, Hengjun and Cui, Long and Wei, Xingguang and Liu, Zhaoyang and Jing, Linglin and Ye, Shenglong and Shao, Jie and others},
  journal={arXiv preprint arXiv:2508.18265},
  year={2025}
}

@article{guo2024large,
  title={Large language model based multi-agents: A survey of progress and challenges},
  author={Guo, Taicheng and Chen, Xiuying and Wang, Yaqi and Chang, Ruidi and Pei, Shichao and Chawla, Nitesh V and Wiest, Olaf and Zhang, Xiangliang},
  journal={arXiv preprint arXiv:2402.01680},
  year={2024}
}

@article{yang2025survey,
  title={A survey of ai agent protocols},
  author={Yang, Yingxuan and Chai, Huacan and Song, Yuanyi and Qi, Siyuan and Wen, Muning and Li, Ning and Liao, Junwei and Hu, Haoyi and Lin, Jianghao and Chang, Gaowei and others},
  journal={arXiv preprint arXiv:2504.16736},
  year={2025}
}

@inproceedings{winata2025worldcuisines,
  title={Worldcuisines: A massive-scale benchmark for multilingual and multicultural visual question answering on global cuisines},
  author={Winata, Genta Indra and Hudi, Frederikus and Irawan, Patrick Amadeus and Anugraha, David and Putri, Rifki Afina and Yutong, Wang and Nohejl, Adam and Prathama, Ubaidillah Ariq and Ousidhoum, Nedjma and Amriani, Afifa and others},
  booktitle={Proceedings of the 2025 Conference of the Nations of the Americas Chapter of the Association for Computational Linguistics: Human Language Technologies (Volume 1: Long Papers)},
  pages={3242--3264},
  year={2025}
}

@article{wang2026personalq,
  title={Personalq: Select, quantize, and serve personalized diffusion models for efficient inference},
  author={Wang, Qirui and Guo, Qi and Sun, Yiding and Yang, Junkai and Zhang, Dongxu and Pang, Shanmin and Guo, Qing},
  journal={arXiv preprint arXiv:2603.22943},
  year={2026}
}
\end{document}